\pdfoutput=1
\documentclass[11pt]{article}

\usepackage[]{EMNLP2023}

\usepackage{booktabs}       %
\usepackage{times}
\usepackage{latexsym}
\usepackage{multirow}
\usepackage{xcolor}
\usepackage{amsmath}
\usepackage{graphicx}
\usepackage{xcolor}
\usepackage{array}
\usepackage{float}
\usepackage[T1]{fontenc}
\usepackage[utf8]{inputenc}
\usepackage{arydshln}
\usepackage{microtype}
\usepackage{titlesec}

\usepackage{inconsolata}

\newcolumntype{C}{>{\centering\arraybackslash}m{1.2cm}}

\title{Visual Storytelling with Question-Answer Plans}

\author{Danyang Liu, Mirella Lapata, Frank Keller \\
  Institute for Language, Cognition and Computation \\
  School of Informatics, University of Edinburgh \\
  10 Crichton Street, Edinburgh EH8 9AB \\
  \texttt{danyang.liu@ed.ac.uk, \{mlap, keller\}@inf.ed.ac.uk}}

\begin{document}

\maketitle
\begin{abstract}
Visual storytelling aims to generate compelling narratives from image sequences. 
% typically employs an encoder-decoder framework.
% keep this sentence or not?
% The encoder transforms the image sequence into an appropriate representation, which the decoder then leverages to create the story.
% Existing models often focus on enhancing the encoder, for example, by using external knowledge sources or advanced graph structures to construct improved input representation.
Existing models often focus on enhancing the representation of the
image sequence, e.g.,~with external knowledge sources or
advanced graph structures. Despite recent progress, the stories are often
repetitive, illogical, and lacking in detail.
% how to attribute the errors?
% indicating a need for more attention to the decoding stage.
% Despite achieving promising results in image description, these models often produce repetitive, illogical, or vague narratives due to limited emphasis on the decoding stage.
To mitigate these issues, we present a novel framework which
integrates visual representations with pretrained language models and
planning. Our model translates the image sequence into a \emph{visual
  prefix}, a sequence of continuous embeddings which language models
can interpret. It also leverages a sequence of question-answer pairs
as a \emph{blueprint plan} for selecting salient visual concepts and
determining how they should be assembled into a narrative.  Automatic
and human evaluation on the VIST benchmark \cite{huang2016visual} 
demonstrates that blueprint-based models generate stories that are
more coherent, interesting, and natural compared to competitive
baselines and state-of-the-art systems. 
 
% We further tested the GPT-3.5 model in our experiments.
% Interestingly, the evaluation results showed its incompatibility with the injection of blueprint signals, which even led to a decrease in the quality of the generated stories, providing an intriguing area for further exploration.

\end{abstract}
\section{Introduction}

\begin{figure}[t]
    \centering
    \includegraphics[width=.48\textwidth]{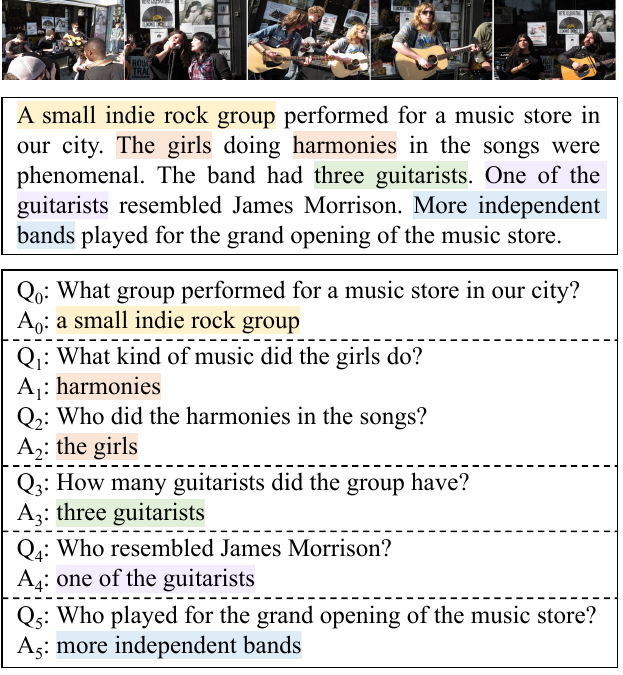}
    \caption{Blueprint annotation for a visual
      story. Color-coded answers are extracted from the gold
      story. Questions are generated by feeding the answers and the
      gold story as context to a pretrained question generator.}
    \label{fig:bp_example}
\end{figure}

Visual storytelling involves narrating an engaging and logically
coherent story based on a sequence of images (see the example in
Figure~\ref{fig:bp_example}). The task lies at the intersection of
natural language processing and computer vision and has recently
attracted increasing interest from both
communities~\cite{wang2022rovist,hsu2021plot,xu2021imagine,chen2021commonsense,hsu2020knowledge,wang2020storytelling,huang2016visual}.
Visual storytelling differs from image captioning, which typically
focuses on generating descriptive text, e.g.,~by identifying and
depicting objects within an image. It requires a deeper understanding
of how images and the events they illustrate relate to each other in
order to create a convincing narrative.

Visual storytelling is commonly modeled as a two-stage process.  The
image sequence is first encoded into a representation which typically
includes image embeddings and detected objects.  Subsequently,
a decoder generates a story token by token based on the encoding of
the image sequence.  Recent work has mainly focused on enhancing the
first stage of the generation process e.g.,~by leveraging external
knowledge sources
\cite{hsu2021plot,chen2021commonsense,hsu2020knowledge,yang2019knowledgeable}.
Advanced representations for image sequences have also been explored,
such as scene graphs \cite{hong2020diverse} and story graphs
\cite{hsu2021plot}. Despite recent progress, these methods struggle to
produce meaningful narratives, are prone to hallucination and
repetition,  often generate vague sentences, and have difficulty
identifying salient visual concepts.

We attribute the lack of story quality to at least two reasons.
Previous work on text-based generation has demonstrated that
\emph{planning} can improve story coherence, allowing to control the
trajectory of events, the characters described and their actions
\cite{yao2019plan,xu-etal-2018-skeleton,rashkin-etal-2020-plotmachines,goldfarb2020content,fan2019strategies,yang2022re3}. However,
planning has not been considered in the context of visual
storytelling, existing models adopt black-box architectures which are
not particularly controllable or interpretable.  Another limitation
concerns the nature of current models which are essentially trained
from scratch, and as a result have limited language modelling and
generalization capabilities (they only see multimodal training samples;
see top of Figure~\ref{fig:bp_example}). Although pretrained
language models \cite{raffel2020exploring, lewis2020bart,
  brown2020language} have been widely adopted for general-purpose
story generation, their potential for visual storytelling remains
unexplored.

%Such models are limited by the available training data and the
%challenges of multimodal training resulting in insufficent language
%modelling and poor generalization.

%Secondly, most existing research still focuses on training Transformer
%models from scratch, resulting in insufficient language modelling and
%poor generalization capabilities of the final model due to the
%difficulty in the multimodal training process.  While large-scale
%pretrained models present a promising solution, their application to
%visual storytelling remains largely unexplored.

%We attribute these limitations primarily to two factors.  Firstly, the
%utilization of planning strategies in the context of visual
%storytelling has yet to be explored.  Evidence suggests that humans
%employ a higher level of planning beyond individual words when
%crafting stories \cite{lebowitz1985story, young2013plans,yao2019plan}.
%However, the black-box nature of deep learning models prevents us from
%observing or explaining the underlying reasoning and planning process
%in text generation.  Moreover, the black box nature also hinders
%controllability as these systems cannot be easily customized to
%individual needs, such as emphasizing a specific entity or character
%in the storytelling process.

In this work we propose an approach to visual storytelling which
integrates pretrained language models with visual representations and
incorporates an intermediate planning step before generating the full
story. Our encoder translates the image sequence into a \textit{visual
  prefix}, a sequence of continuous embeddings which language models
can interpret.  Following \citet{narayan2022conditional}, we represent
plans as a sequence of question-answer pairs, called \emph{blueprints},
which serve as a proxy for content selection (i.e.,~what to say) and
planning (i.e.,~in what order). Blueprints are loosely related to the
Question-under-Discussion (QUD) theory of discourse
\cite{larsson2002issue,roberts2012information,de2020towards}, which
posits that text structure can be analyzed by identifying
\emph{implicit} questions raised and answered by subsequent spans of
text. We augment visual storytelling training data with story blueprints (see Figure~\ref{fig:bp_example}), which we
obtain automatically thanks to state-of-the-art question generation
technology.

We fine-tune pretrained language models to generate blueprints from
image sequences \emph{and} the stories based on them. We showcase two
types of storytelling models, which vary in how the planning mechanism
is implemented. A \emph{top-down} model generates the blueprint first
and then continues to generate the corresponding story in one go,
whereas an \emph{integrated} model interleaves planning with text
generation rather than determining a plan in advance; generation is
iteratively conditioned on the image input, the blueprint and the
story generated so far. Experiments on the VIST benchmark
\cite{huang2016visual} show that blueprint-based models generate more
coherent, interesting, and human-like stories compared to the state of
the art and large language models (LLMs) like GPT-3.5, according to automatic and human evaluation.

%seamlessly integrated into pretrained language models, which
%effectively maps the visual semantic space to a textual space that
%large language models can interpret.  More specifically, the visual
%encoder translates the image sequence into a \textit{visual prefix} —
%a sequence of continuous embeddings that can be interpreted by the
%language model. 

%We fine-tune pretrained language models to generate
%blueprint-story pairs from an input image sequence, as opposed to
%generating the story only.

% The contributions of our work include:
% \begin{itemize}
%     \item A visual encoder designed for seamless integration into pretrained language models, enabling an effective mapping from visual semantic space to textual space.
%     \item A \textit{Blueprint}-based planning strategy that uses a sequence of question-answer pairs as intermediate story plans for improved visual concept selection and story quality.
%     \item Empirical evidence from experiments on the VIST dataset, showcasing the superior performance of our \textit{BP-VIST} model compared to state-of-the-art models and large language models like GPT-3.5.
% \end{itemize}

%%% Local Variables: 
%%% mode: latex
%%% TeX-master:  "emnlp2023"
%%% End:

\section{Related Work}
\paragraph{Visual Storytelling}
 \citet{huang2016visual} introduced visual storytelling as a vehicle
 for developing AI tools with human-like understanding of grounded
 event structure and linguistic abilities that go beyond descriptive
 language. While earlier work \cite{gonzalez2018contextualize,
   kim2018glac} employed simple encoder-decoder architectures (using
 CNNs to extract visual features and RNNs to generate text), more
 recent methods
 \cite{xu2021imagine,chen2021commonsense,hsu2020knowledge,yang2019knowledgeable}
 leverage external resources (e.g.,~ConceptNet) as a way of instilling
 commonsense reasoning skills.  Sometimes, scene graphs are also used
 to model relations between objects
 \cite{lu2016visual,hong2020diverse,wang2020storytelling}.  To our
 knowledge, none of these approaches make use of plan-based decoding.
 \citet{hsu2021plot} construct a graph representing the image sequence
 (based on training data and external resources) and identify the
 highest scoring path as the best storyline encapsulated therein.  The
 storyline can be viewed as a form of planning, however, on the
 encoder side.

Most existing approaches
\cite{xu2021imagine,hsu2020knowledge,wang2020storytelling,yang2019knowledgeable}
train Transformer models from scratch, with the exception of
\citet{chen2021commonsense}, who employ a vanilla BART model as a
baseline without task-specific adaptation.  In contrast, our work
leverages the language modeling and generalization capabilities of
pretrained language models for visual storytelling.
%% todo

\paragraph{Planning and Generation}
In the domain of automatic story generation, planning has been
effective at capturing the content and structure of stories. The
generation process is often decomposed into two stages, namely planning
an outline and then elaborating on it, e.g., by filling in specific
details of a story. Plans have been represented as a sequence of
event or phrase keywords
\cite{yao2019plan,xu-etal-2018-skeleton,rashkin-etal-2020-plotmachines},
character actions \cite{liu2020character}, plot structures
\cite{goldfarb2020content}, and more elaborate descriptions
including details about the setting of the story, its characters, and
main plot points \cite{yang2022re3}.

%present a system that utilizes a
%plot-generation language model and an ensemble of rescoring models to
%learn and implement Aristotle's Poetics-based aspects of good
%story-writing for effective content planning in story generation.

The idea of having a separate planning stage has also been explored
for other text generation tasks including summarization
\cite{narayan-etal-2021-planning,liu-chen-2021-controllable} and
data-to-text generation
\cite{moryossef-etal-2019-step,puduppully2022data}. Our work is
closest to \citet{narayan2022conditional} who propose the use of
question-answer pairs as intermediate plans for summarization.
However, their approach is designed for descriptive text. Our work
extends their framework to a multimodal setting, where the input
consists of image sequences, and the output are narratives
characterized by more abstract and figurative language.

%%% Local Variables: 
%%% mode: latex
%%% TeX-master:  "emnlp2023"
%%% End:

\section{Blueprint-based Visual Storytelling}

Let $\mathit{I}$ represent a sequence of~$k$ images, denoted as
$\{v_1, v_2 ..., v_k\}$.  Given this input, our goal is to generate a
blueprint plan~$\mathit{B}$ (i.e., an ordered set of question-answer
pairs) and a story~$\mathit{S}$ based on it.  Most generation datasets
do not include blueprint annotations, and visual storytelling is no
exception. We first describe how we automatically obtain $\{I_i, B_i,
S_i\}_{i=1}^{N}$ training data samples (Section~\ref{bp-creation}),
and then introduce our story generation models (Section~\ref{model}).

\subsection{Blueprint Annotation}
\label{bp-creation}
Let $\{I_i, S_i\}_{i=1}^{N}$ denote a dataset consisting of pairs of
image sequences and their corresponding stories. We automatically
create blueprint~$B_i$ based on story~$S_i$ using state-of-the-art
question generators
\cite{mromero2021t5-base-finetuned-question-generation-ap,raffel2020exploring},
coupled with a filtering procedure to remove repetitions and
ill-formed questions.

%\paragraph{Question-Answer Generation}
The generation of question-answer pairs involves two steps, namely
answer extraction and question generation. In the context of
storytelling, capturing key events is crucial for a compelling
narrative. While noun phrases and named entities are commonly
recognized as significant content units in other tasks such as
summarization \cite{narayan2022conditional,deutsch2023incorporating},
verb phrases also play a vital role in conveying story dynamics,
actions, and relationships
\cite{trabasso1989logical,eisenberg-finlayson-2017-simpler,liu2020character}. Therefore,
in addition to named entities and noun phrases, we also extract verb
phrases as answer candidates using the \href{https://spacy.io/}{spaCy} library.

We then generate questions for answer candidates with a~T5 model
\cite{raffel2020exploring,mromero2021t5-base-finetuned-question-generation-ap}
fine-tuned on the SQuAD reading comprehension dataset
\cite{rajpurkar2018know}. The answer and the story are provided as context
to predict the corresponding question. We decontextualize stories by
replacing pronouns with their corresponding head mentions, using a
state-of-the-art coreference resolution model
\cite{dobrovolskii-2021-word}.

%\paragraph{Question-Answer Filtering}
Question-answer pairs are subsequently filtered to eliminate noise
which is unavoidable due to the  automatic preprocessing steps
mentioned above.  We thus remove any question-answer pairs where the
answer is already present in the question. We  also employ a round-trip
consistency check \cite{alberti-etal-2019-synthetic} which discards
questions if they yield answers different from those used to generate
them.

%as a refinement step.  Specifically, we use a T5-based
%question-answering model pretrained on the SQuAD dataset.  This model
%takes the generated question and the story as context and predicts the
%corresponding answer. 
%We filter out the question-answer pairs if the answers generated by
%the model differ from the ones used to generate the questions. 
%This filtering process helps maintain consistency and reliability in
%the generated blueprint. 
% add a figure here to show?
% This filtering process not only helps maintain consistency but also detects grammatically incorrect answers, as shown in the Figure.

\subsection{Blueprint Models}
\label{model}

Our approach leverages the generation capabilities of pre-trained
sequence-to-sequence models. As our backbone model, we
employ BART-base \cite{lewis2020bart} which has been fine-tuned for
text generation. We adapt this model to our visual storytelling task
in two ways. Aside from enabling the generation of blueprints, we
convert the image sequence to a \emph{visual prefix} which the
pretrained language model can interpret. The pretrained language model is prompted with this prefix to generate the blueprint, and eventually the story.

\begin{figure}[t]
    \centering
    \includegraphics[width=.48\textwidth]{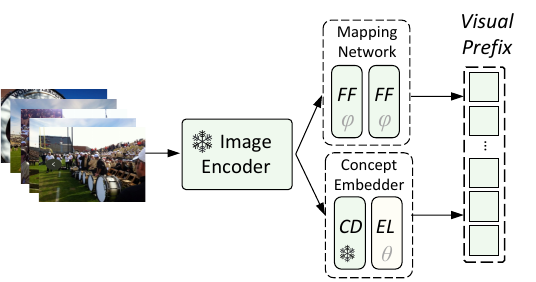}
    \caption{Visual prefix construction. The pretrained image encoder
      and concept detector are frozen. \textit{FF} refers to a
      feed-forward layer, \textit{CD} and \textit{EL} denote a concept
      detector and embedding layer, respectively.}
    \label{fig:vla}
\end{figure}

\begin{figure*}[t]
    \centering
    \includegraphics[width=.98\textwidth]{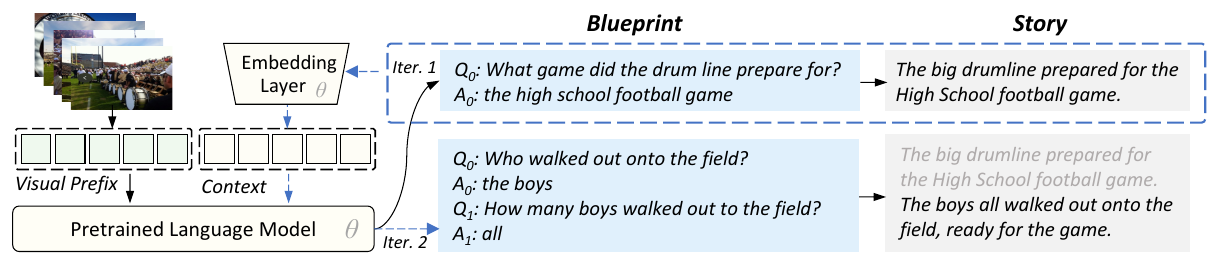}
    \caption{Iterative blueprint model: in the first iteration, the
      embedding layer uses \textit{<START>} as context; in the second iteration, the context includes the blueprint and previously generated story (enclosed by blue
      dashed line). In contrast, the top-down model takes only the
      visual prefix as input and predicts a global blueprint and
      corresponding story in one go. For details on the visual
      prefix, see Figure~\ref{fig:vla}.}
    \label{fig:arch}
\end{figure*}

\paragraph{Visual Prefix Construction} 
Our model needs to grasp what the image sequence is about, e.g., the
depicted objects, actions, and their associations.  Drawing
inspiration from recent advances in vision and language research
\cite{mokady2021clipcap,tsimpoukelli2021multimodal,alayrac2022flamingo,zhai2022lit,liu2023llava,huang2023language},
we translate the input sequence of images into a sequence of
continuous embeddings, aka a visual prefix (see
Figure~\ref{fig:vla}).

Following previous work
\cite{wang2018no,xu2021imagine,chen2021commonsense}, we use
ResNet-152 \cite{he2016deep} to extract visual features from images.
We next employ a lightweight linear mapping network that consists of
a series of feedforward layers, denoted as~$F_\phi$, to map image
features to $k$~visual clues:
\begin{equation}
p_1, \ldots, p_k=F_\phi\left(\operatorname{ResNet}\left(v_1, v_2, \ldots, v_k\right)\right)
\end{equation}
where $k$ is the image sequence length, and each visual clue $p_i$ has the same dimensionality as a token embedding of the pretrained language model.

To further instil world knowledge in our visual prefix, we employ a concept detector. The latter identifies specific objects within images, but also actions, scenes, and attributes.
For each image~$v_i$, we retain the $K$ concepts $\{c_i^1, c_i^2, ..., c_i^K\}$
with the highest confidence score:
\begin{equation}
c_i^1, \ldots, c_i^K=\operatorname{Concept}\left(\operatorname{ResNet}\left(v_i\right)\right)
\end{equation}
Concepts for each image are then concatenated with a $\langle\mathit{SEP}\rangle$
token and serve as input to the embedding layer of the pretrained
model. The visual clues and concept embeddings are concatenated to
form the visual prefix~$\mathit{V}$.  The image encoder and the
concept detector remain frozen during the training phase. Only the
parameters in the mapping network~$F_\phi$ are updated (see
Figure~\ref{fig:vla}).

%Our objective is to develop a method for generating blueprints and
%corresponding stories, denoted as $(B, S)$, based on an input image
%sequence $\mathit{I}$.  To achieve this, we introduce the
%\textit{VLA-BP} approach, which utilizes pretrained language models to
%harness their powerful language modeling capabilities.  Additionally,
%we propose a V-L adaptor that represents the image sequence as a
%visual prefix, enabling a pretrained language model to interpret the
%visual signals when prompted with this specific prefix.  In our
%method, we propose two models: an End2end blueprint model, which
%generates the story in a single pass based on a global blueprint, and
%an Iterative blueprint model, which employs a local blueprint for each
%subsequent sentence and generates the story incrementally.  An
%illustration of the model architecture is provided in
%Figure~\ref{fig:arch}.

\paragraph{Top-down Planning}
This model takes image sequence~$\mathit{I}$ as input and 
generates  blueprint~$\mathit{B}$ and story~$\mathit{S}$ in one go.
More precisely, during decoding, the model first generates the
blueprint, which then serves as a prompt guiding subsequent story
generation.  Our training objective maximizes the log-likelihood of
the joint distribution:
\begin{equation}
\max _{\theta, \phi} \sum_{i=1}^N \log p_{\theta,\phi}\left(\mathit{B}_i,\mathit{S}_i \mid\mathit{I}_i\right)
\end{equation}
where $(\mathit{B},\mathit{S})$ refers to the concatenation of 
blueprint~$\mathit{B}$ and story~$\mathit{S}$. $\theta$~represents the
parameters of the pretrained language model, $\phi$~are the parameters
of the mapping visual network~$F_\phi$, and $N$~denotes the size of
the dataset.

We introduce special tokens \textit{Story:} and \textit{Plan:}
preceding the story and blueprint, respectively. In experiments,
our blueprints consist of answer-question pairs $\{a_1, q_1, \ldots,
a_m, q_m\}$ (rather than question-answer pairs).  We place the answer
before its question to encourage the model to zoom in on salient
visual concepts depicted in the image sequence.  This ordering is
intuitive for our storytelling task: We first decide on what the story
is about and then elaborate on key concepts.  Incidentally,
\citet{narayan2022conditional}  also find that generating  the answer
before the question performs better for their summarization
tasks. Finally, the model is trained with the standard maximum
likelihood objective to generate the joint target.

%This ordering also aligns with the creation of the
%blueprint annotation, as described in Section \ref{bp-creation}.
%Besides, \citep{narayan2022conditional} suggested that generating the
%question prior to the answer could lead to inferior performance.

\paragraph{Iterative Planning}
This  model employs an incremental generation strategy to
create the story.  Rather than generating in one step a global blueprint and the
story, planning and generation are interleaved.  At each
time step, the iterative model considers the image sequence
\emph{and} context from previous steps, including the blueprint and
story generated so far.  We gradually construct the blueprint and its
corresponding story sentence-by-sentence; our planning is informed by
generation and vice versa, which we argue should be mutually
beneficial (they are conditioned on each other).

Let $\mathit{S}=\{s_1, s_2, \ldots, s_k\}$, denote a target story and
$\mathit{B}=\{b_1, b_2, \ldots, b_k\}$ its blueprint where
$s_i$~represents the $i$-th sentence in the story, and $b_i$ its
associated blueprint.  Each~$b_i$ consists of answer-question
pairs, denoted as $\{a_1^{i}, q_1^{i}, \ldots, a_{l(i)}^{i},
q_{l(i)}^{i}\}$, where $l_{(i)}$ is the number of pairs in the $i$-th
blueprint.  The training objective for the iterative model is defined
as follows:
\begin{equation}
\max _{\theta, \phi} \sum_{j=1}^N \sum_{i=1}^k \log p_{\theta,\phi}\left(b_{i+1},s_{i+1}|b_{1:i},s_{1:i},\mathit{I}_j \right)
\end{equation}
where $b_{1:i}$ and $s_{1:i}$ refer to the blueprint and sentences generated so far, from step $1$ to step $i$.

At each time step~$i$, the encoder takes image sequence~$\mathit{I}$
as input, and the decoder takes the context (i.e., blueprint and
sentences generated so far $\{b_1, b_2, \ldots, b_{i}; s_1, s_2,
\ldots, s_i\}$) as a prompt to predict the next blueprint~$b_{i+1}$
and sentence $s_{i+1}$.  Therefore, the iterative model is trained on
samples $\{\mathit{I}, (b_{1:i}, s_{1:i}), b_{i+1}, s_{i+1}\}$.  We
prefix $(b_{1:i}, s_{1:i})$, $b_{i+1}$, and $s_{i+1}$ with
\textit{Context:}, \textit{Plan:}, and \textit{Next Sentence}, respectively.  To handle the first time step, we
introduce special token $\langle\mathit{START}\rangle$ as context to predict~$b_1$ and~$s_1$.  We also use $\langle\mathit{END}\rangle$ to indicate the completion of
an iteration (see Figure~\ref{fig:arch} for an illustration).  It is
important to note that $(b_{1:i}, s_{1:i})$ are masked out when
computing the loss because they serve as prompts to the decoder.  We
want to avoid the model repeatedly predicting and overly optimizing
the blueprints and sentences that appear at the beginning of the
output.

%%% Local Variables: 
%%% mode: latex
%%% TeX-master:  "emnlp2023"
%%% End:

\section{Experimental Setting}

\begin{table}[tbp]
\centering
\resizebox{.45\textwidth}{!}{%
\begin{tabular}{lr}
%Name & Value \\
\toprule
Length of image sequences & 5.0 \\
Number of sentences in the story &  5.0\\
Number of tokens per story & 52.3 \\
Number of QA pairs per story & 11.1 \\
Number of tokens per QA pair & 10.3\\
Number of tokens per story plus QA pair & 166.2\\
\bottomrule
\end{tabular}%
}
\caption{VIST dataset Statistics (average values).}
\label{tab:statistic}
\end{table}

\subsection{Dataset}
We performed experiments on the widely used VIST dataset
\cite{huang2016visual}, which contains 10,117~Flickr albums and
210,819~unique photos.  Each training sample consists
of~$k\operatorname{=}5$ images and a corresponding story
of~$k\operatorname{=}5$ sentences.  As described in
Section~\ref{bp-creation}, we augment each story with an
automatically generated blueprint.
% The statistics about the enhanced dataset are shown in Figure.

\subsection{Implementation Details}
Our models are built on top of BART-base \cite{lewis2020bart} and
finetuned with a learning rate of~\mbox{3e-5}, batch size of~64, and
warm-up ratio of~0.05.  We select the best checkpoint on the
validation set using a QA-based metric which quantifies the extent to
which the output story follows its blueprint (see
Section~\ref{sec:auto-eval}).  During inference, we employ beam search
(size~5). For our visual prefix, we employed the \href{https://clarifai.com/clarifai/main/models/general-image-detection?tab=overview}{Clarifai Concept Detector} which was trained on a
dataset containing 9,098~concepts and 20 million images (multiple
concepts are assigned to each image), and is integrated with the
InceptionV2 architecture \cite{szegedy2016rethinking} .

\subsection{Comparison Systems}
We compared our models against several baselines and state-of-the-art
visual storytelling models. These included a vanilla BART-base model
  with the same encoder and visual prefix as ours but no planning
  (\textbf{VP-BART}; it generates the story directly in an
  autoregressive manner without the blueprint).  \textbf{KG-Story}
  \cite{hsu2020knowledge} predicts a set of words representative of
  the image sequence, enriches them using external knowledge graphs,
  and generates stories based on the enriched word set.
  \textbf{PR-VIST} \cite{hsu2021plot} is a state-of-the-art model which
  constructs a graph representing the
  relations between elements in the image sequence, identifies the
  best storyline captured therein, and proceeds to generate a story
  based on it.  The process of constructing the story graph can be
  viewed as a form of planning.  Along similar lines,
  \citet{chen2021commonsense} build a common sense knowledge graph
  capturing concepts in the image sequence, and use \textbf{MCSM}, a
  Maximal Clique Selection Module to identify which ones to write a
  story about. They use BART-large to generate the story based on 
  selected concepts (and image features).

We also compared against an LLM that generates
stories via prompting. We provide
\href{https://platform.openai.com/docs/models/gpt-3-5}{GPT-3.5} with
a visual prefix, namely the concepts identified in the image sequence,
and a prompt which explains how to create the blueprint and generate
the story together with examples (in-context learning). Details on the
prompt can be found in Appendix~\ref{sec:appendix}.

\subsection{Automatic Evaluation}
\label{sec:auto-eval}

We evaluated our stories using BLEU, ROUGE, METEOR, and
CIDER, mainly to compare to previous work.  Several studies 
\cite{hsu2022learning,hsu2021plot,hsu2020knowledge,hu2020makes,yang2019knowledgeable,modi2019steep}
have demonstrated the inadequacy of
lexical matching metrics:
they correlate poorly with human judgments, and not do effectively
measure the semantic similarity to human-written stories or the lexical
richness of the generated stories.
%With these caveats in mind, we
%report results using BLEU, Rouge, METEOR, and CIDER, mainly to compare
%with with previous studies.

%\subsection{Story-Specific Metrics}

We further employ \textit{story-specific} metrics to assess story
quality aspects such as diversity, fluency, naturalness, and
grounding. Specifically, we use two types of trigram repetition
metrics \cite{yao2019plan,goldfarb-tarrant-etal-2020-content}.
\emph{Intra-story repetition} is a fluency metric, it measures the
proportion of trigrams
repeated  \textit{within} a story. 
\emph{Inter-story repetition} examines trigram repetition
\textit{across} stories.  This metric evaluates diversity, high
intra-story repetition suggests that the model tends to generate the
same story even when conditioned on different image sequences.
We also use MAUVE \cite{pillutla-etal:mauve:neurips2021} to measure
the \emph{naturalness} of the generated stories.  MAUVE is a recently
introduced automatic metric for open-ended generation which has high
correlation with human judgements. It computes the similarity of the
distribution of human-written text and machine-generated text.

To quantify the extent to which the generated story is
\emph{grounded}, i.e.,~whether it accurately represents the content of
the image sequence, we measure \emph{concept precision} and \emph{recall}.
Precision measures the number of words in the generated story that align with the detected concept set, while  recall  assesses the number of words in the detected concept set that are present in the generated story.

Finally, for our own models we also evaluate whether the generated
stories are \emph{faithful} to their blueprint.  
Drawing inspiration from recent studies on summary evaluation
\cite{deutsch2021towards,fabbri2022qafacteval}, we measure how well the
generated story answers questions from the predicted blueprint. We
utilize a RoBERTa-based \cite{liu2019roberta} QA model finetuned on the
SQuAD dataset.

\section{Results}

\begin{table*}[t]
\centering
\resizebox{\textwidth}{!}{
\begin{tabular}{l||cc|cr|r||c||rccr}
\toprule
\multicolumn{1}{l||}{\multirow{2}{*}{Model}} &
\multicolumn{2}{c|}{Repetition ($\downarrow$)} &
\multicolumn{2}{c|}{Grounding ($\uparrow$)} &
\multicolumn{1}{c||}{\multirow{2}{*}{MAUVE} \raisebox{-1ex}[0pt]{($\uparrow$)}} &
\multicolumn{1}{c||}{\multirow{2}{*}{Faithful ($\uparrow$) }} &
\multicolumn{4}{c}{N-gram-based Metrics ($\uparrow$)} \\ 
\multicolumn{1}{c||}{} & Intra & Inter & Precis. & Recall & \multicolumn{1}{c||}{} & \multicolumn{1}{c||}{} & B-4 & RLSum & METEOR & \multicolumn{1}{l}{CIDER} \\ 
\hline
KG-Story & 1.03 & 88.72 & 4.55 & 3.46 & 3.86& --- & 9.8 & 27.3 & 32.3 & \textbf{7.9} \\
PR-VIST & 1.19 & 83.80 & 3.76 & 3.28 & 2.31& ---& 7.5 & 26.1 & 31.4 & 7.6 \\
MCSM & 2.85  & 77.48  & 5.12 & 5.89 & 11.01 &  --- & 8.1 & 27.7 & 31.4 & 7.6 \\
\hline
VP-BART & 0.22 & 83.70 & 4.31 & 3.23  & 11.31 & --- & 8.6 & 26.6 & 31.0 & 6.8 \\
~~~ + BP (top-down) & \textbf{0.08} & 81.51 & 5.17 & \textbf{11.56} & 8.32 & 44.73 & \textbf{9.9} & \textbf{28.5} & \textbf{33.6} & 7.2 \\
~~~ + BP (iterative) & 0.29 & \textbf{72.70} & \textbf{5.22} & 3.59 & \textbf{28.25} & \textbf{51.66} & 7.0 & 26.1 & 30.3 & 5.5\\ 
\quad+ BP (gold) & 0.12 & 18.61 & 6.81 & 2.97  & 52.24 & --- & 29.4 & 52.0 & 58.4 & 36.3 \\ \hline
GPT-3.5 & 0.47 & 40.61 & 10.80 & 7.90  & 2.30 &  --- &  5.0 & 24.4 & 27.3 & 1.9 \\
GPT-3.5 + BP  & 1.52 & 31.19 & 14.70 & 10.30 & 2.10 & 34.56 &  4.2 & 23.3 & 25.1 & 2.3\\
\bottomrule
\end{tabular}%
}
\caption{Automatic evaluation results. We report intra- and
  inter-story trigram Repetition (lower is better), precision and
  recall for concept grounding, MAUVE, Faithfulness, and a suite of
  commonly used metrics which rely on lexical similarity between
  system stories and references. Best results are highlighted in bold
  font.}
\label{tab:auto-results}
\end{table*}

Our results are summarized in Table~\ref{tab:auto-results}. The first
block presents the performance of state-of-the-art storytelling
systems. The second block presents variants of our approach: a vanilla
BART model, enhanced with a visual prefix (VP), and two blueprint
models which vary in the way plans are generated, i.e., in a top-down
fashion or iteratively. The third block contains GPT-3.5 models with
(+BP) and without blueprints.

\paragraph{Pretrained Language Models Produce Better Stories}
We observe that models based on pretrained language models (i.e.,~our models and MCSM,
outperform models trained from scratch (i.e., KG-Story and PR-VIST) in
terms of trigram-repetition scores and MAUVE.  This indicates that we
can maintain strong language modeling capabilities while enabling pretrained language models to process visual signals effectively.

\paragraph{The Visual Prefix is an Effective Interface between Image
  and Text} MCSM is the only existing model that utilizes a pretrained
language model for visual storytelling.  However, our baseline model
(VP-BART) demonstrates superior performance in most story-specific
metrics.  Remarkably, this is achieved using a smaller pretrained
model (BART-base, 140M parameters); MCSM is built on top of BART-large
(400M parameters).  This highlights the effectiveness of our visual
prefix, indicating it successfully translates the image sequence into
a space that BART can understand.

%\begin{figure}[t]
%    \centering
%    \includegraphics[width=.48\textwidth]{img/fig3.pdf}
%    \caption{Automatic evaluation results of \textit{VLA-BP} with silver blueprints as input, indicating the upper-bound performance. The left panel represents \textit{VLA-IterBP}, while the right panel represents \textit{VLA-GloBP}.}
%    \label{fig:upper}
%\end{figure}

\paragraph{Blueprint Models are Most Grounded}
Our models outperform comparison systems in terms of concept
grounding.  This confirms that an intermediate planning step allows
the model to effectively select salient concepts based on the visual
prefix.  The top-down model in particular achieves the
highest concept grounding recall, it stays close to the image
sequence, accurately describing the information conveyed therein. The
higher lexical matching scores further support this observation. The
iterative blueprint model achieves the best concept grounding
precision (excluding GPT-3.5 models) which in turn suggests that the
stories generated by this model  exhibit a stronger grounding to the
images with fewer hallucinations.

\paragraph{The Iterative Model Generates Most Natural and Faithful Stories}
Despite not achieving the highest scores in lexical matching metrics,
the iterative blueprint model stands out in terms of MAUVE evaluation.
Compared to other models, it generates more natural stories, closer to
those written by humans.  This finding suggests that humans might
employ a similar iterative planning strategy, at least for the short
stories considered here; they construct a narrative gradually rather
than a global plan which they subsequently convert into a story.

With regard to faithfulness, we observe that both blueprint models
achieve scores higher than 40\%, indicating effective translation of
blueprints into stories.  Notably, the iterative model performs best
in terms of faithfulness, which suggests that translating the entire
global blueprint into a story is more challenging, whereas breaking
down planning into individual steps is more effective. To get an idea
of the upper bound performance for blueprint models, we ran the
top-down model with silver standard blueprints extracted from the
human-written stories (see row +BP~(gold) in
Table~\ref{tab:auto-results}). As can be seen, the MAUVE score jumps
to 52.24, edging closer to human-written stories (their MAUVE score
is~69.6).  This further supports our hypothesis that  our model
successfully leverages the blueprints and retains the information
captured in them.

\begin{table*}[t]
\centering
\resizebox{\textwidth}{!}{%
\begin{tabular}{l||C|C|r||C|C|r||r|C|r}
\toprule
\multirow{2}{*}{Choices(\%)} & \multicolumn{3}{c||}{Iterative vs. PR-VIST} & \multicolumn{3}{c||}{Iterative vs. VP-BART} & \multicolumn{3}{c}{Iterative vs. Human} \\ %\cline{2-10} 
 & Win & Lose & Tie & Win & Lose & Tie & Win & Lose & Tie \\ \hline
Fluency & \textbf{47.0} & 37.6 & 15.4 & 40.5 & \textbf{42.9} & 16.6 & 10.9 & \textbf{84.3} & 4.8 \\
Coherence & \textbf{56.0} & 24.8 & 19.2 & \underline{41.2} & \underline{\textbf{41.6}} & 17.2 & 15.2 & \textbf{75.7} & 9.1 \\
Interestingness & \textbf{70.5} & 19.2 & 10.3 & \textbf{55.5} & 38.1 & 6.4 & 23.0 & \textbf{70.0} & 7.0 \\
Grounding & \textbf{50.9} & 40.6 & 8.5 & \textbf{45.1} & 41.7 & 13.2 & 7.8 & \textbf{79.6} & 12.6 \\ \hline
Overall & \textbf{58.1} & 25.6 & 16.2 & \textbf{42.6} & 41.5 & 15.9 & 12.6 & \textbf{80.0} & 7.4 \\ \bottomrule
\end{tabular}%
}
\caption{Human evaluation results. Raters provide 
  pairwise story preferences  in terms of 
  fluency, coherence, interestingness, grounding, and overall. VP-BART
  is a BART model enhanced with a visual prefix (VP) but no planning;
  Iterative is our best blueprint model and PR-VIST is a
  state-of-the-art visual storytelling model. We report the percentage
  of times the Iterative model Wins, Loses or is in a Tie with a 
  comparison system. Unless \underline{underlined}, differences between systems are statistically significant ($p<0.05$; using  the Wilcoxon signed-rank test).}
\label{tab:human-results}
\end{table*}

\begin{figure*}[t]
    \centering
    \includegraphics[width=\textwidth]{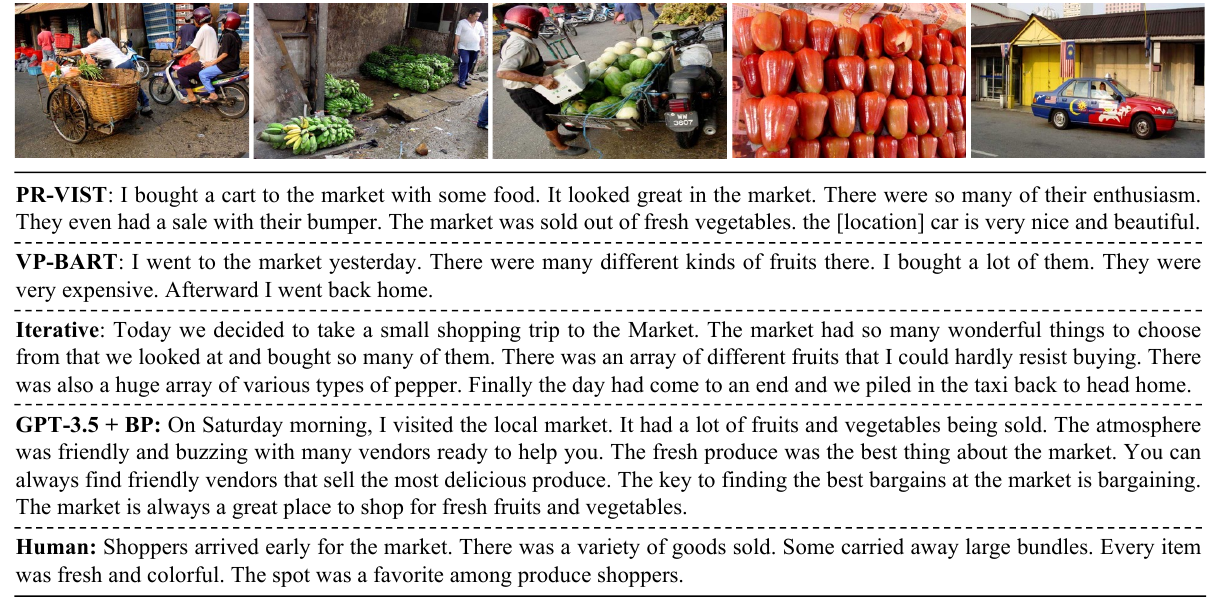}
    \caption{Examples of system output and human-written story for an image sequence.}
    \label{fig:case}
\end{figure*}

\paragraph{GPT-3.5 Struggles with Blueprints}
We also compared our approach to GPT-3.5, which we adapted to our task
with in-context learning. A GPT-3.5 model enhanced with blueprints
performs well at concept grounding, i.e., it generates stories which
talk about what is depicted in the image. However, these stories are
neither human-like (see the very low MAUVE score) nor faithful to the
intermediate blueprints (in fact they  are~10\% less faithful compared to our
iterative model). This suggests that GPT-3.5 tends to ignore the plan,
despite being explicitly prompted with blueprints.

\section{Human Evaluation}
\label{sec:humaneval}

We conducted a judgment elicitation study to further evaluate the
stories generated by our models.  Specifically, we compared the best
performing blueprint model (iterative) and three other systems:
(a)~{PR-VIST}, which represents the current planning-based state of
the art; (b)~{VP-BART}, our proposed model without blueprints; and
(c)~ground truth stories written by humans.  Raters were shown an
image sequence, alongside two stories and were asked to provide
pairwise references along the following dimensions:
\textit{Relevance}, \textit{Fluency}, \textit{Coherence},
\textit{Interestingness}, and \textit{Overall}. The full instructions
are given in Appendix~\ref{sec:appendix}. We recruited nine
native English speakers and elicited preferences for
100~stories (three judgments per story).

Our human evaluation results are summarized in
Table~\ref{tab:human-results}. The iterative
blueprint model outperforms {PR-VIST} across 
metrics. Our participants perceive VP-BART stories  as marginally more fluent and coherent compared to those created by
the iterative model (even though they prefer iterative stories overall). This discrepancy is likely due to the
generation process introduced by the iterative model  which requires the
decoder to produce a mix of questions, answers, and corresponding
sentences, deviating from the traditional BART pretraining pattern.
This added complexity might result in
minor grammatical errors and pose challenges for coherence, given that
story generation is broken down into separate steps instead of being a
continuous process.  Nonetheless, the coherence scores are fairly
close.

The blueprint model excels in terms of interestingness and grounding,
indicating its effectiveness in creating engaging and memorable
stories.  Our model's superior grounding performance aligns with our hypothesis that blueprints serve not only as a planning strategy but also as a visual concept selector.
This is due to the way blueprints are structured (as answer-question pairs), which explicitly
forces the model to first identify salient visual concepts and then
generate questions based on them.

Figure~\ref{fig:case} shows example stories created by the models used
in our human evaluation study and \mbox{GPT-3.5+BP}.  The story
generated by the iterative model is coherent, rich in detail, and
fluent.  VP-BART generates a grounded and accurate story
without hallucination and semantic errors. However, it is a relatively
plain narrative, offering limited detail about the market or the
experience of the characters.  Compared to \mbox{GPT-3.5+BP}, the
iterative model's story follows the image sequence more closely,
mentioning details like \textit{an array of different fruits} and
\textit{various types of pepper}, which significantly enhances
storytelling.

\section{Controllable Generation}
\label{sec:app-contro} In this section we showcase how the blueprint plan allows us to control the content and length of model output without additional mechanisms or training.  
%We showcase examples of faithful story generation and long story generation by leveraging the inherent controllability of our blueprint method.
%\paragraph{Faithful Story Generation}

For example, in cases where the generated story  contains entities which do not appear in the image sequence, it is possible to refine the story generation process, mitigating hallucinations. Specifically, we apply a filtering step which removes non-grounded entities (and corresponding QA pair) from the blueprint before generating the story. We  consider as  \emph{non-grounded} any blueprint entity which is 
not included in the output of the concept detector (see Section~\ref{model}).

\begin{figure}[t]
    \centering
    \includegraphics[width=.47\textwidth]{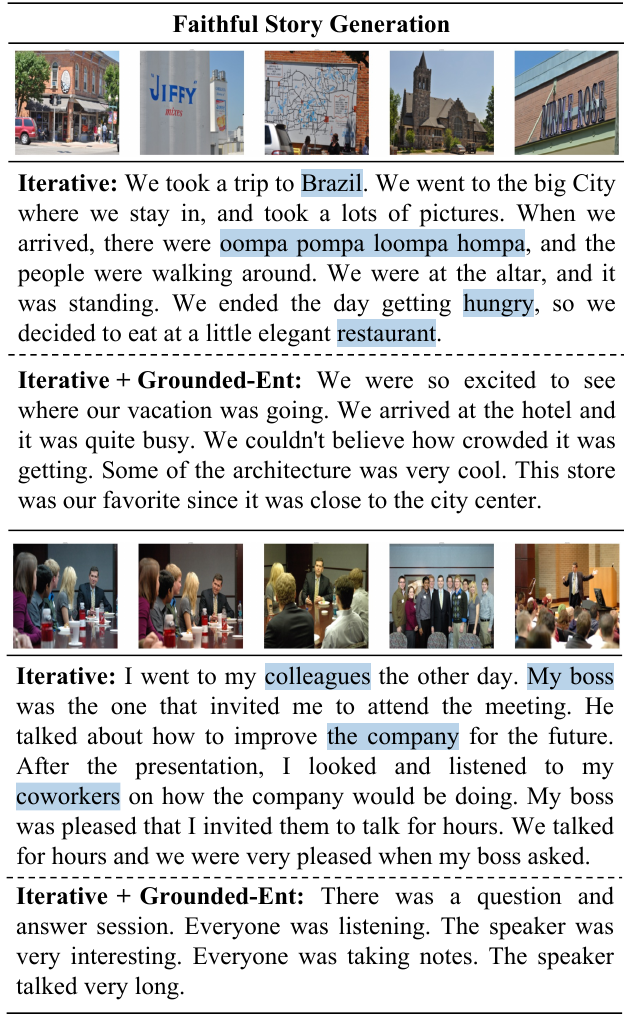}
    \caption{Comparison of stories generated by original blueprint and refined blueprint models. Hallucinated words are highlighted in blue.}
    \label{fig:ground}
\end{figure}

Figure~\ref{fig:ground} shows how this refinement approach can be used to adjust the model's output.
In the first example, we observe that the story generated with a refined blueprint effectively avoids hallucinations (highlighted in blue) and is overall more faithful. However, it is important to note that imagination plays a crucial role in crafting an engaging story, especially when the  image sequence provides limited information. Therefore, employing the refinement method may result in shorter and less detailed stories, as illustrated in the second example. While the refined blueprint successfully eliminates all hallucinated entities,  the resulting story appears plain and lacks depth. Our blueprint method seems to strike the right balance between accurate and captivating story generation, prioritizing  faithfulness to the image sequence and creativity in storytelling.

\begin{figure}[t]
    \centering
    \includegraphics[width=.47\textwidth]{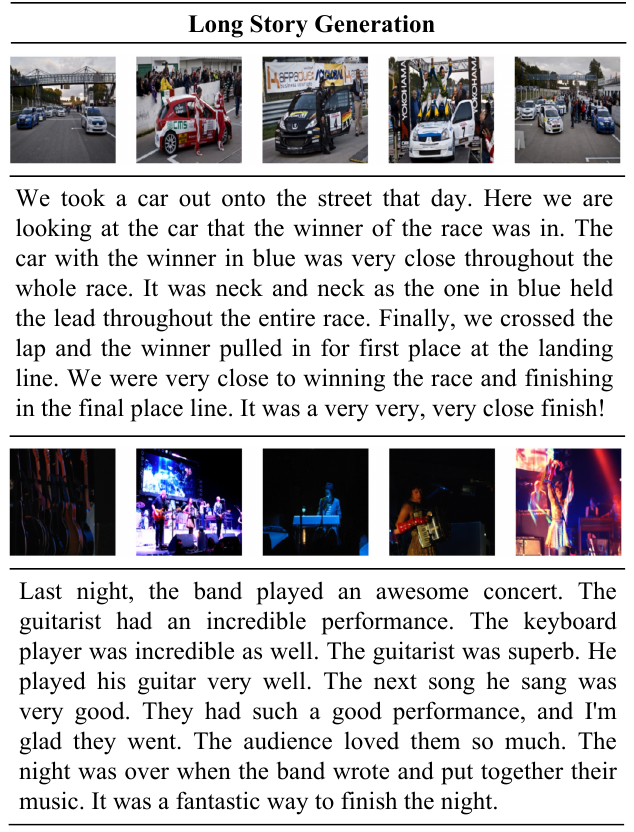}
    \caption{Stories generated within 10 iterations.}
    \label{fig:long}
\end{figure}

%\paragraph{Long Story Generation}
Most visual storytelling systems generate \mbox{5-sentence} stories, following the predefined story structure of the VIST
dataset \cite{huang2016visual}. Nevertheless, our iterative blueprint
model can flexibly modulate the length of the story by controlling
the number of iterative steps, thereby overcoming the conventional
sentence limitation.  Figure~\ref{fig:long} presents stories generated
by this model with a maximum of 10~iterations. Despite the increased length, the stories maintain coherence and are engaging.

\section{Conclusion}
In this work, we have introduced a novel approach to visual
storytelling which integrates visual representations with pretrained
language models and a blueprint-based planning method for story
generation.  Blueprint models leverage a sequence of question-answer
pairs as intermediate plans, enabling better selection of salient
concepts from the image sequence and guiding the construction of the
final narrative.  Specifically, we have showcased two model variants:
a top-down model which relies on a global plan, and an iterative
model, which interleaves planning with sentence generation.  Our
experiments have shown that blueprint models excel in concept
grounding and their ability to create human-like stories.  Additionally,
they are controllable: Blueprints can be made shorter or longer and their
details can be refined (e.g., by emphasising specific entities or
characters), thus enabling human-in-the-loop and personalized
storytelling. We showcase examples of controllability in Section~\ref{sec:app-contro}.
In the future, we would like to explore visual storytelling with grounded characters and entities, as well as tackle the generation of more complex narratives, such as long-form stories.

\paragraph{Acknowledgments}
 The authors  gratefully acknowledge the support of the
UK Engineering and Physical Sciences Research
Council (grant EP/W002876/1). Liu was
supported  by the UKRI Centre for Doctoral
Training in Natural Language Processing, funded
by the UKRI (grant EP/S022481/1) and the University of Edinburgh.

%%% Local Variables: 
%%% mode: latex
%%% TeX-master:  "emnlp2023"
%%% End:

% \clearpage
\section*{Limitations}
While our proposed model demonstrates effective story generation, it has certain limitations. Firstly, the grounding relation between the visual concepts and the corresponding text may not always be clear, leading to potential ambiguity in the generated stories. Furthermore, the model can sometimes suffer from hallucinations due to  falsely detected visual concepts.

It is worth noting that our model was built on top of BART-base \cite{lewis2020bart}. It would be beneficial to investigate the performance of larger models, as they could potentially  enhance the quality of the planning component and overall storytelling capability. 

\section*{Ethics Statement}

\paragraph{Large Language Models} This paper uses large pretrained  language models, which have been shown to be subject to a variety of biases, to occasionally generate toxic language, and to hallucinate content.  Model output used for the human evaluation study  (Section~\ref{sec:humaneval}) was screened by the authors for harmful content. 

\paragraph{Experimental Participants} The departmental ethics panel judged our human evaluation study to be exempt from ethical approval, as all participants were employees of the University of X, and as such were protected by employment law. Participants were paid at the standard hourly rate for tutors and demonstrators at the university.

% \section*{Acknowledgements}

\bibliography{custom,anthology}
\bibliographystyle{acl_natbib}

\appendix
\clearpage

\begin{figure*}[t]
    \centering
    \includegraphics[width=.98\textwidth]{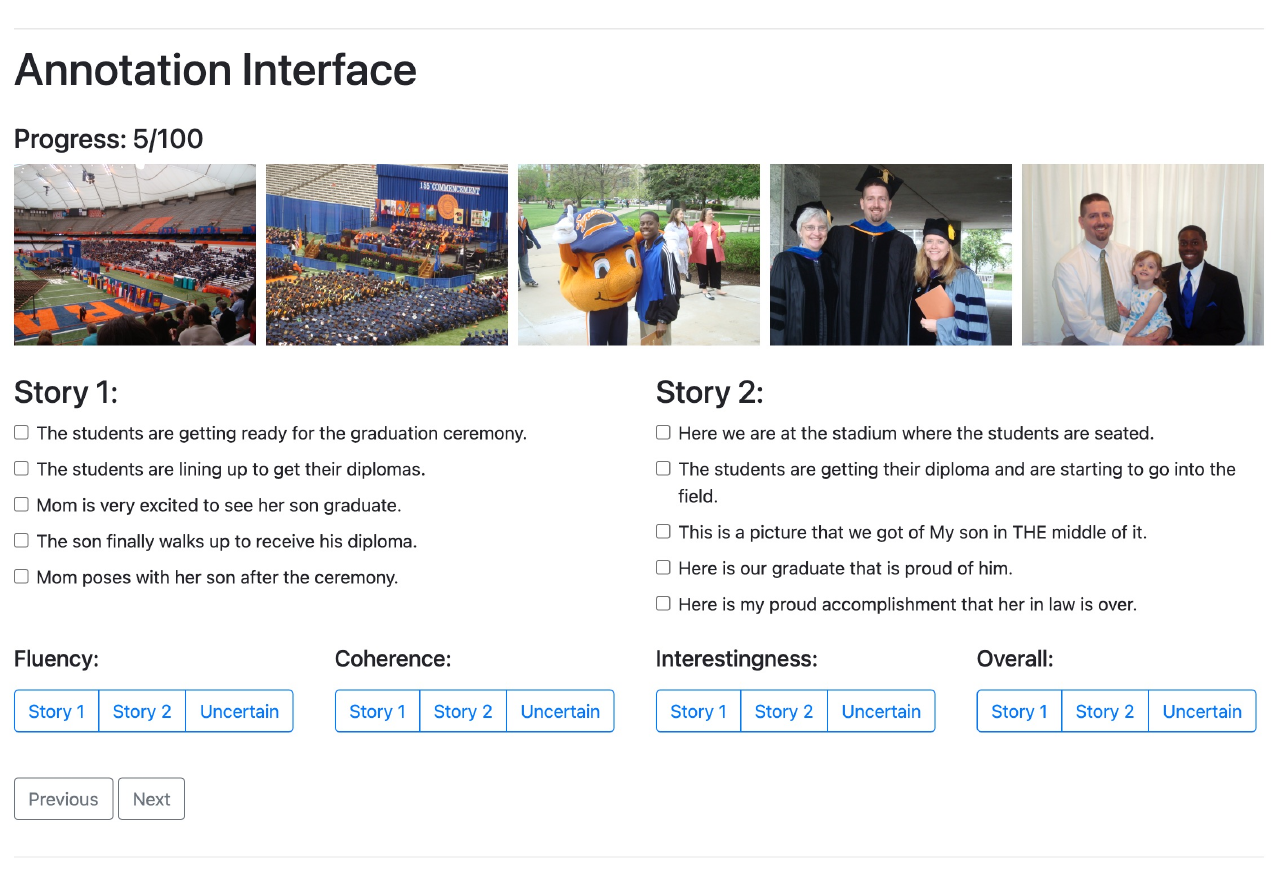}
    \caption{Screenshot of the annotation interface.}
    \label{fig:interface}
\end{figure*}

\section{GPT-3.5 Experimental Setting}
\label{sec:appendix}

% \begin{table}[H]
%     \centering
%     \begin{tabular}{p{7.3cm}}
%     \toprule
%         \multicolumn{1}{c}{Prompt (w/o Blueprint)} \\ \midrule
%         I want you to act as a visual storyteller by creating a story based on a given sequence of images. For each image, I'll provide the key concepts. The concepts from different images will be separated by <SEP>. \textbackslash n Concepts: ..., \textbackslash n Story: ...\\ \midrule
%         \multicolumn{1}{c}{Prompt (w/ Blueprint)} \\  \midrule
%         I'd like you to act as a visual storyteller by creating a story, based on a given sequence of images. For each image, I'll provide the key concepts, which will be separated by '<SEP>'. Your task is to first generate a series of question-answer pairs for each image as part of the planning process, and then use these pairs to create the final story. \textbackslash n Concepts: ..., \textbackslash n Plan: ..., \textbackslash n Story: ... \\
%         \bottomrule
%     \end{tabular}
%     \caption{Prompts used in our GPT-3.5 experiments. Examples for
%       in-context learning examples are not shown in the table for the
%       sake of brevity.}
%     \label{tab:prompt}
% \end{table}

We designed the following prompts for GPT-3.5:
\paragraph{Prompt (w/o Blueprint):} \textit{I want you to act as a visual storyteller by creating a story based on a given sequence of images. For each image, I'll provide the key concepts. The concepts from different images will be separated by <SEP>. \textbackslash n Concepts: ..., \textbackslash n Story: ..., ..., \textbackslash n Concepts: ..., \textbackslash n Story:}
\paragraph{Prompt (w/ Blueprint):} \textit{I'd like you to act as a visual storyteller by creating a story based on a given sequence of images. For each image, I'll provide the key concepts separated by <SEP>. Your task is to first generate a series of question-answer pairs for each image as part of the planning process, and then use these pairs to create the final story. \textbackslash n Concepts: ..., \textbackslash n Plan: ..., \textbackslash n Story: ..., ..., \textbackslash n Concepts: ..., \textbackslash n Plan: ..., \textbackslash n Story:}

In-context learning examples in the prompts are not shown for brevity.
To make the best of in-context learning, we employed
max-shot learning, while adhering to the token limit of 4,096.

%  lists the prompts used for GPT-3.5 experiments. 
% We use max-shot in-context learning within the length limitation (max 4,096 tokens), specifically, 8 examples in prompt for in-context learning for GPT3.5 w/o BP and 7 examples for GPT3.5 w/ BP.

\section{Human Evaluation Study}

As mentioned in Section~\ref{sec:humaneval} we conducted a judgment elicitation study to further evaluate the
stories generated by our models.  Specifically, we compared the best
performing blueprint model (iterative) and three other systems:
(a)~{PR-VIST}, which represents the current planning-based state of
the art; (b)~{VP-BART}, our proposed model without blueprints; and
(c)~ground truth stories written by humans. Raters were shown an
image sequence, alongside two stories and were asked to provide
pairwise preferences along various dimensions of story quality. We
describe below our evaluation procedure and reproduce the instructions
given to our raters. 

\subsection{Evaluation Procedure}
We initially conducted a pilot study, based on which we devised our
instructions.  Subsequently, 100 image sequences were randomly
selected from the test set, leading to a total of 300 pairwise
comparisons.  We employed the expertise of 9 native speakers who
triple-annotated the 300 pairwise comparisons, resulting in a total of
900 judgments.

\subsection{Experimental Instructions}
Human raters were asked to compare and evaluate stories generated by
different systems using pairwise judgments. The evaluation focused on
the following dimensions of story quality: \textit{Relevance},
\textit{Fluency}, \textit{Coherence}, \textit{Interestingness}, and an
\textit{Overall} judgment. We provide their definitions below. 

\paragraph{Relevance}
Relevance captures whether the sentences in the stories relate to the input images.
For each sentence, if the sentence accurately describes the content of a specific image (i.e., not imagining something that does not exist in the image), it will be marked as relevance. Otherwise, if the sentence does not correspond to an image or is too vague, it will be not marked as relevance.
\paragraph{Fluency}
Fluency evaluates the grammatical correctness of the text. A story is fluent if it has few or no grammatical errors and is easy to understand.
\paragraph{Coherence}
Coherence assesses whether the story makes sense. A coherent story flows well, the sentences are related, and logically connected. In contrast, an incoherent story would be more or less incomprehensible, without any logical connection between its sentences.
\paragraph{Interestingness}
This evaluates whether the story contains unique, possibly unexpected elements. For example, a memorable storyline. Below are examples of a dull story and an interesting story.
\paragraph{Overall}
Taking all the aforementioned criteria into consideration, the annotators selected their preferred story for the given set of five images.

\subsection{Annotation Interface}
We designed an annotation interface using Python
Flask. Figure~\ref{fig:interface} shows a screenshot of the interface.

\end{document}